# Convolutional Sparse Kernel Network for Unsupervised Medical Image Analysis

Euijoon Ahn[a], Ashnil Kumar[a], Michael Fulham[b,c], Dagan Feng[a,d], and Jinman Kim[a,*]

[a] School of Computer Science, University of Sydney, NSW, Australia.
[b] Department of Molecular Imaging, Royal Prince Alfred Hospital, Camperdown, NSW, Australia.
[c] Sydney Medical School, University of Sydney, Camperdown, NSW, Australia
[d] Med-X Research Institute, Shanghai Jiao Tong University, China.

E. Ahn (E-mail: eahn4614@uni.sydney.edu.au)
A. Kumar (E-mail: ashnil.kumar@sydney.edu.au)
M. Fulham (E-mail: michael.fulham@sydney.edu.au)
D. Fang (E-mail: dagan.feng@sydney.edu.au)
J. Kim[*] (E-mail: Jinman.kim@sydney.edu.au)

[*] Corresponding Author

Declarations of interest: none

## Abstract

The availability of large-scale annotated image datasets and recent advances in supervised deep learning methods enable the end-to-end derivation of representative image features that can impact a variety of image analysis problems. Such supervised approaches, however, are difficult to implement in the medical domain where large volumes of labelled data are difficult to obtain due to the complexity of manual annotation and inter- and intra-observer variability in label assignment. We propose a new convolutional sparse kernel network (CSKN), which is a hierarchical unsupervised feature learning framework that addresses the challenge of learning representative visual features in medical image analysis domains where there is a lack of annotated training data. Our framework has three contributions: (i) We extend kernel learning to identify and represent invariant features across image sub-patches in an unsupervised manner. (ii) We initialise our kernel learning with a layer-wise pre-training scheme that leverages the sparsity inherent in medical images to extract initial discriminative features. (iii) We adapt a multi-scale spatial pyramid pooling (SPP) framework to capture subtle geometric differences between learned visual features. We evaluated our framework in medical image retrieval and classification on three public datasets. Our results show that our CSKN had better accuracy when compared to other conventional unsupervised methods and comparable accuracy to methods that used state-of-the-art supervised convolutional neural networks (CNNs). Our findings indicate that our unsupervised CSKN provides an opportunity to leverage unannotated big data in medical imaging repositories.

## 1. Introduction

Medical imaging is now ubiquitous in modern healthcare because it provides invaluable data for patient diagnosis and management. Most current medical imaging data are digital and stored in vast imaging repositories. These repositories or archives provide new opportunities for evidence-based and computer-aided diagnosis, physician training and biomedical research (Litjens et al., 2017) (Kumar et al., 2013). Computer-aided diagnosis systems (CADs) can automatically analyse, categorise, and retrieve images, by relating low-level image features to high-level semantic concepts or expert domain knowledge using machine learning approaches. These supervised approaches use prior knowledge derived from labelled training data and approaches, such as convolutional neural networks (CNNs) have produced impressive results in natural (photographic) image classification (Simonyan and Zisserman, 2014),(He et al., 2016),(Szegedy et al., 2015). CNNs learn image features in a hierarchical fashion. Each deeper layer of the network learns a representation of the image data that is high-level and semantically more meaningful. For example, in image classification, the learned features can be a class-specific representation (Le, 2013) to enable better discrimination between different image classes (Simonyan and Zisserman, 2014),(He et al., 2016). These CNNs require a large number of annotated training images, e.g., ImageNet with over 1 million natural images. Such large image datasets are scarce in the medical domain because the images can be difficult to interpret and image labelling / annotation is costly, tedious, slow, and subject to clinician inter- and intra-observer variability (Shin et al., 2013).

Transfer learning was introduced to address the lack of large amounts of labelled medical image data through a model that was pre-trained on a different domain, e.g., natural images as a generic feature extractor, or through using a relatively small dataset of medical images to optimise a pre-trained model from a different domain, i.e., fine-tuning (Kumar et al., 2017),(Tajbakhsh et al., 2016),(Shin et al., 2016),(Bi et al., 2017). Unfortunately, both approaches rely on general image features derived from a different domain and they are unable to capture the high-level semantic features, which are most relevant to a specific dataset. As a result, they have inferior accuracy when compared to approaches that learn image features directly from large, specific annotated data. An alternative approach is to use unsupervised feature learning algorithms to build features from unlabelled data, which then allows unannotated image archives to be used (Lee et al., 2006),(Hinton et al., 2006),(Nair and Hinton, 2010),(Le, 2013),(Erhan et al., 2010),(Romero et al., 2016). Many of these methods, however, have only shown strong performance in learning low-level features such as 'lines' or 'edges' (Lee et al., 2006),(Hinton et al., 2006),(Nair and Hinton, 2010),(Le, 2013). Many unsupervised methods were used to pre-train a model that was later coupled to a supervised learning stage, i.e., the unsupervised component was used as a pre-training phase to derive useful priors that acted as an initialisation point for the supervised fine-tuning (Erhan et al., 2010),(Romero et al., 2016). Thus the onus was on the supervised phase to learn high-level and semantically meaningful image features. Our aim was to derive a framework that enables learning semantically meaningful image features in a completely unsupervised fashion.

1.1 Related work

Many unsupervised feature learning approaches are based on sparse coding (Lee et al., 2006), sparse auto-encoders (Hinton et al., 2006), and Restricted Boltzmann Machines (RBMs) (Nair and Hinton, 2010) and are limited to learning and extracting low-level features. Only a few methods, such as the stacked sparse autoencoder (SSAE) reported by Le et al, where the SSAE pre-trained a model was coupled to supervised deep learning (i.e., fine tuning), have been able to extract semantic high-level features. Highly non-linear and non-parametric models are crucial to unsupervised feature learning algorithms (Song et al., 2018). Kernel learning is a natural approach to derive non-linear models via a similarity function in a reproducing kernel Hilbert space (RKHS) (Zhuang et al., 2011). Machine learning techniques have been adapted to a RKHS and have improved performance in object recognition and clustering (Thiagarajan et al., 2014). Recently, deep learning architectures have been used for kernel learning (Mairal et al., 2014; Song et al., 2018) with state-of-the-art performance in natural image classification (Mairal et al., 2014) and retrieval (Paulin et al., 2015). These architectures learned data representations in a RKHS and a non-linear hierarchical manner, but they are prone to overfitting (the learning cost function often gets stuck in local minima) when the training data are small.

The concept of sparsity is widely used in computer vision and has proven effective in image compression (Skodras et al., 2001), denoising (Buades et al., 2005), tomographic reconstruction, segmentation (Ahn et al., 2015; Ahn et al., 2017; Zhang et al., 2012), and classification (Ahn et al., 2016; Jiang et al., 2011). Sparsity can be used to derive compact and optimal representations of image data, where trivial information or parameters can be ignored without compromising image quality or characteristics (Leahy and Byrne, 2000). Recently, sparsity-based CNNs have been also applied to supervised deep learning approaches to reduce the number of parameters in the architecture (Graham et al., 2018; Graham and van der Maaten, 2017; Liu et al., 2015; Liu et al., 2018). (Liu et al., 2015) were able to reduce 90% of the parameters in dense CNNs that then provided a marked in improvement in computational speed. It has been shown that in medical image data, feature representations have an intrinsic sparse structure under certain fixed bases (e.g., Fourier) (Li et al., 2012; Lustig et al., 2007). Lustig et al., (2007) improved temporal resolution of magnetic resonance (MR) imaging by adding a sparsity constraint; this step then allowed the development of a number of novel CADs in cardiac and brain imaging. This intrinsic sparsity often comes in two complementary forms (Willmore and Tolhurst, 2001): population and lifetime sparsity. Population sparsity refers to the activation of small subsets of the bases (i.e., a sparse set of the population) to encode different information; only a small subset of the coding outputs (feature maps or bases) are active for any given stimulus (input images), and different subsets are active for different stimuli. This ensures that the activation of different bases is a discriminator for different image data. In contrast, lifetime sparsity refers to the short frequency of activation of bases for different inputs (i.e., each base has a sparse lifetime); different bases are active very rarely and each activation has a high response. This ensures that the strong rare activations are indicators for higher degrees of information (the higher the information, the higher the entropy) in the underlying image data. Motivated by these findings, we suggest that incorporating sparsity into layerwise unsupervised pre-training will allow the extraction of more discriminative features for medical image data. Sparse pyramid pooling (SPP) can represent the spatial layout of image features by partitioning the image into multi-level regions and aggregating local features (Lazebnik et al., 2006). SPP has been successfully applied to image classification (Yang et al., 2009), (Wang et al., 2010) and object detection (Van de Sande et al., 2011).

### 1.2. Contribution

We have designed an unsupervised deep learning framework to learn semantic high-level features from unlabelled medical images, which we refer to as the Convolutional Sparse Kernel Network (CSKN), to address the challenge of learning representative visual features in medical image analysis where there is a lack of annotated training data. Our CSKN derives a kernel space for modelling image similarity that is constrained by the inherent image sparsity and the local geometric properties of distinct classes. Since our CSKN represents images in a kernel space it can derive features that are highly non-linear in a non-parametric manner, which is crucial for unsupervised feature learning (Song et al., 2018). Furthermore, the derived kernel space depicts a stronger discriminative local semantic representation of the imaging data by ignoring trivial or redundant parameters (Leahy and Byrne, 2000). The main contributions of our work are:

1) a new approach to characterise medical images by combining kernel learning and CNNs to learn invariant local features in a hierarchical manner;

2) an unsupervised convolutional sparse feature learning algorithm that effectively learns initial discriminative features in a RKHS;

3) initialising the weights of a kernel network that can then be pre-trained in a layer-wise fashion and,

4) incorporating a SPP framework that provides more discriminative and geometrically invariant local feature representations of medical image data.

The remainder of paper is organised as follows: a) materials used in this paper and proposed framework are introduced in Section 2; b) details of the implementation and experimental setup are described in Section 3; c) evaluation of the framework in comparison to different methods is provided in Section 4; d) we discuss our findings, limitations and future work in Section 5 and, e) we summarise the work in Section 6.

## 2. Material and methods

### 2.1 Datasets

### 2.1.1 IRMA X-ray dataset

The Image Retrieval in Medical Application (IRMA) dataset comprises 14,410 gray-scale X-ray images with 193 hierarchical classes (Lehmann et al., 2004),(Lehmann et al., 2003). The IRMA dataset contains images with irregular contrast, brightness, and artifacts, with high intra-class variability and inter-class similarity. We used the standard pre-defined training set of 12677 images and test set of 1733 images (Lehmann et al., 2003). The images were annotated according to the IRMA coding system with four different axes, as described by (Lehmann et al., 2003): 1) a technical code that describes imaging modality, 2) a directional code for imaging orientation, 3) an anatomical code for body region examined, and 4) a biological code for biological system examined. Fig 1 illustrates a sample X-ray image and the corresponding labels from the IRMA code.

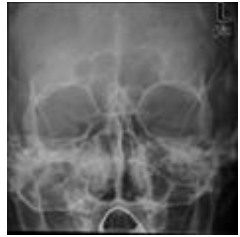

| **IRMA Code** | 1121-420-212-700 |
| --- | --- |
| **Technical Code** | X-ray, Plain radiography, Overview Image |
| **Directional Code** | Other orientation, occipitofrontal |
| **Anatomical Code** | Facial cranium, eye area |
| **Biological Code** | Musculoskeletal system |

Fig 1. A sample X-ray image (Face) and the corresponding labels from IRMA code.

### 2.1.2 ImageCLEF dataset

We used the medical Subfigure Classification dataset used in the Image Conference and Labs of the Evaluation Forum (ImageCLEF) 2016 competition (García Seco de Herrera et al., 2016; Villegas et al., 2016). We used the standard pre-defined training set of 6776 images and test set of 4166 images from 30 different image modalities. Ground truth annotations are available for both image datasets. While a multitude of different types of images have been collected to assist in the development of more advanced CADs, the labelling of the collated image data remains problematic (Müller et al., 2007),(Müller et al., 2008),(Müller et al., 2010),(Müller et al., 2012). In cases where appropriate labels are absent, automatic identification of the imaging modality is an initial important step because the semantics and content of an image can vary greatly depending on the modality.

### 2.1.3 ISIC dataset

We used the skin diseases classification dataset from the International Skin Imaging Collaboration (ISIC) 2017 competition (Codella et al., 2018). The dataset is a clinical dataset and contains 2000 training images and 600 test images with 3 different diagnoses of skin lesions (benign nevus, seborrheic keratosis, and melanoma). Ground annotations were obtained from expert clinicians as well as pathology reports. The clinical dermoscopy images in this dataset have complex and diverse image characteristics for the important challenge of recognising different skin conditions.

## 2.2 Methods

### 2.2.1 Overview of the CSKN framework

Fig 2 is an overview of our CSKN framework. We first used a kernel map to represent the local features of medical image data. Then, as a pre-training step, we learned convolutional sparse features in a RKHS as a starting point of convolutional kernel learning. We then learned a multi-layer kernel network in a feedforward manner. Finally, we applied SPP to extract a final image representation that captures subtle and discriminative geometric variations.

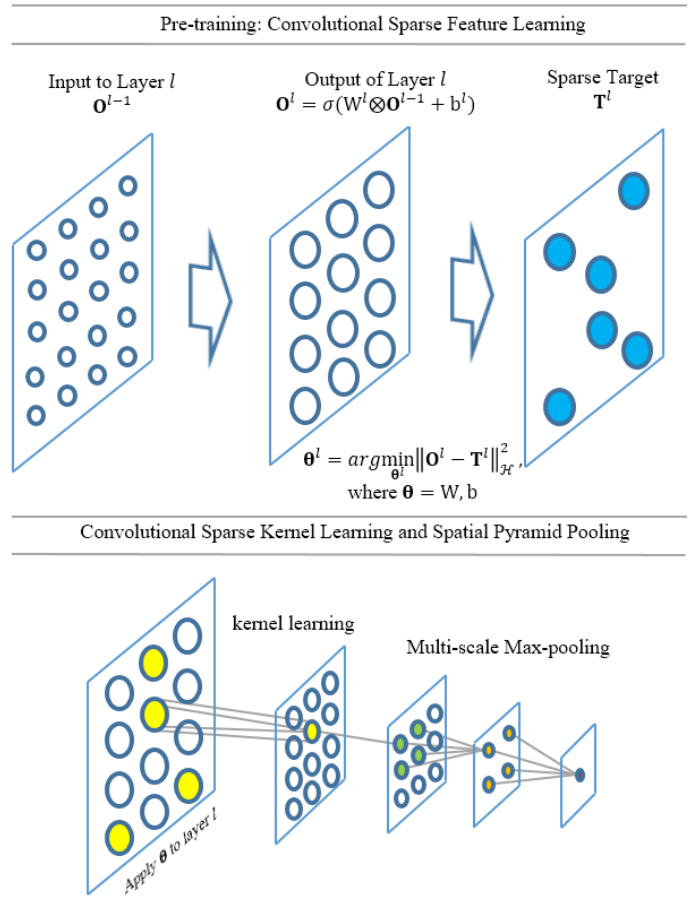

Fig 2. Our proposed framework.

### 2.2.2 Background: convolutional neural networks (CNNs)

CNN layers generally have: 1) convolutional layers to learn weights (i.e., filters) that can be used to extract features from the input; 2) a linear operation followed by a pointwise non-linearity such as the sigmoid function or rectified linear units and, 3) pooling layers to aggregate features that are in spatial proximity (down-sampling the data in the process). The output of single layer CNN can be represented as:

$$f(\mathbf{O}) = pool_p\big(\sigma(\mathbf{W} \otimes \mathbf{O} + \mathbf{b})\big), \quad (1)$$

where $\mathbf{O}$ is the input feature map, $\sigma(\cdot)$ is the pointwise non-linear function, and $\boldsymbol{\theta} = \{\mathbf{W}, \mathbf{b}\}$ are the set of parameters (i.e., weights and biases). The *pool* function denotes a down-sampling operation and $p$ is the size of pooling region. The symbol $\otimes$ indicates the linear convolution. When a convolutional layer is dense and unstructured, it is called "fully connected". For example, the well-established AlexNet (Krizhevsky et al., 2012) CNN has 8 trainable layers comprising five convolutional layers followed by three fully connected layers. Training such a CNN, however, is challenging because of the number of hyperparameters that need to be carefully tuned. Some major hyperparameters include the size of learnable filters, the number of layers, the number of outputs per layer, and the size of the down-sampling factor. Sub-optimal hyperparameter choice leads to overfitting and an inability to derive optimal high-level semantic image features. Some supervised CNNs have exploited unsupervised layerwise pre-training schemes to render better generalisation of image data (Le, 2013),(Romero et al., 2016). The pre-training acts as a form of regularisation which minimises variance and restricts the range of the parameter values for subsequent supervised training (Erhan et al., 2010). Layerwise unsupervised pre-training allows all the available unlabelled image data to be used to pre-train the network's local parameters, which potentially provides a good initialisation point for further supervised training.

### 2.2.3 Combining kernel learning with CNNs

Our CSKNs have the classic hierarchical architecture of CNNs but use kernel maps to represent image features. A kernel map is used to understand the local geometry of the image data by modelling invariance (Mairal et al., 2014). We suggest that kernels coupled with a hierarchical architecture allow the effective learning of image features without a reliance on labels. The architecture of a two-layer CSKN is shown in Fig 3. Let us consider two image patches $O$ and $O'$ of an image of size $m \times m$ ($m = 200$ in this paper), with $\Omega$ being a set of pixel coordinates ($\Omega = \{1, \ldots m\}^2$). Given the locations $z$ and $z'$ in $\Omega$, let $s_z \in O$ and $s'_z \in O'$ be sub-patches of the image feature map, we define a single layer convolutional kernel network as follows (Mairal et al., 2014):

$$K(O, O') = \sum_{z,z' \in \Omega} \|s_z\|_{\mathcal{H}} \|s'_z\|_{\mathcal{H}} e^{-\frac{1}{2\beta^2}\|z-z'\|_2^2} e^{-\frac{1}{2\alpha^2}\|\tilde{s}_z - \tilde{s}'_z\|_{\mathcal{H}}^2}. \quad (2)$$

where $\|\cdot\|_{\mathcal{H}}$ denotes the Hilbertian norm. The kernel $K$ is a positive definitive kernel that consists of a sum of pairwise comparisons between image features of sub-patches. The term $\|s_z\|_{\mathcal{H}} \|s'_z\|_{\mathcal{H}}$ acts to emphasise the spatial and feature similarity (captured by the exponential terms) for non-small intensity-valued patches. The term $e^{-\frac{1}{2\beta^2}\|z-z'\|_2^2}$ captures spatial distance between $z$ and $z'$, and the term $e^{-\frac{1}{2\alpha^2}\|\tilde{s}_z - \tilde{s}'_z\|_{\mathcal{H}}^2}$ measures the feature similarity between sub-patches. These two terms work in conjunction with the Hilbertian norm terms to create a kernel that gives larger values for patches that are close in both space and intensity. We used two different types of input feature maps:

1) Patch map: the sub-patch $s_z$ is an image sub-patch size $b \times b$ centred at $z$. The sub-patch $s_z$ is simply $\mathbb{R}^{b \times b}$ and $\tilde{s}_z$ denotes a contrast-normalised version of the sub-patch.

2) Gradient map: the sub-patch $s_z$ is the two-dimensional gradient of the image at pixel $z$, which is computed with first-order differences along each dimension. In this formulation, $\|s_z\|_{\mathcal{H}}$ is the gradient intensity and $\tilde{s}_z$ denotes its orientation defined as an angle with $[\cos\theta, \sin\theta]$ (Bo et al., 2010). When the input data is in a compact set ($\mathbb{R}^d, d \leq 2$), Equation (2) can be approximated by uniform sampling over a large enough set; the term $e^{-\frac{1}{2\beta^2}\|z-z'\|_2^2}$ indicates a spatial kernel and $e^{-\frac{1}{2\alpha^2}\|\tilde{s}_z - \tilde{s}'_z\|_{\mathcal{H}}^2}$ denotes the gradient map.

The coefficients $\beta$ and $\alpha$ are smoothing Gaussian kernel parameters that control spatial distances between $z$ and $z'$ and the feature closeness between $\tilde{s}_z$ and $\tilde{s}'_z$ in the Hilbert space, respectively. The corresponding kernel map is formalised as a weighted match kernel between all sub-patches from training samples that defines a feature representation of the image.

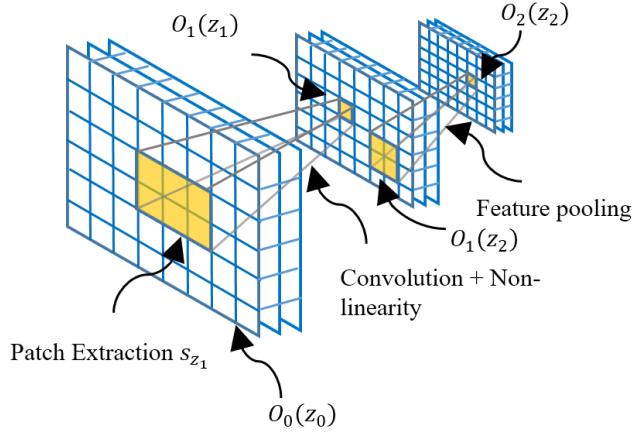

Fig 3. A two-layer CSKN; each layer is a weighted match kernel between all sub-patches of the previous layer.

2.2.4    Unsupervised feature learning via CSKNs

Match kernels are expensive to compute when the input data has high dimensionality ($\mathbb{R}^d, d > 2$). The computational complexity also grows quadratically with increasing sample sizes. To prevent the curse of dimensionality, we used a fast approximation approach with finite-dimensional embedding proposed by (Mairal et al., 2014). For all $u \in \Omega_1$ and $z \in \Omega$:

$$K(O, O') \approx \sum_{u \in \Omega_1} g(u; O)^T g(u; O') \quad (3)$$

$$g(u; O) \coloneqq \sum_{z \in \Omega} e^{-\frac{1}{2\beta^2}\|u-z\|_2^2} h(z; O) \quad (4)$$

$$h(z; O) \coloneqq \|s_z\|_2 \left[ \sqrt{b_i} e^{-\frac{1}{\alpha^2}\|W_i - \bar{s}_z\|_2^2} \right]_{i=1}^{n_1}, \quad (5)$$

where $\Omega_1$ is a subset of $\Omega$, $n_1$ denotes number of filters, and b and W are learned parameters. This operation can be considered to be similar to a spatial convolution of the feature map followed by a pointwise non-linearity. Since $K(O, O')$ is a sum of the match kernel terms, we can learn to approximate the kernel using training data. The parameters b and W are learned at the sub-patch level by solving an optimisation problem:

$$\min_{W_i, b_i} \sum_{c=1}^{n} \left( e^{-\frac{\|\bar{s}_c - \bar{s}'_c\|_2^2}{2\alpha^2}} - \sum_{i=1}^{p} b_i e^{-\frac{\|W_i - \bar{s}_c\|_2^2}{\alpha^2}} e^{-\frac{\|W_i - \bar{s}'_c\|_2^2}{\alpha^2}} \right). \quad (6)$$

We randomly selected 400,000 pairs of sub-patches from the training data and used the standard Limited memory Broyden Fletcher Goldfarb Shanno with Bounds (L-BFGS-B) (Byrd et al., 1995) optimiser to solve Equation (6) (Mairal et al., 2014). The L-BFGS-B requires less parameters and can be superior to the conjugate gradient (CG) or stochastic gradient descent (SGD) in many applications such as image classification (Ngiam et al., 2011).

2.2.5    Initialisation of CSKN via layerwise unsupervised pre-training with sparsity

We formulated a layerwise unsupervised feature learning algorithm that efficiently enforces population and lifetime sparsity (EPLS) in a RKHS. Our approach learns convolutional sparse features in a RKHS, in contrast to Romero et al's (2015) original EPLS algorithm that learns sparse features from decomposed raw image patches. The convolutional sparse features learned in the unified feature space are often more discriminative and therefore allow us to build more class-specific representations (Thiagarajan et al., 2014). Furthermore, the convolutional features learned by our method preserve the relationships between neighbourhood pixels so as to learn local structures and reduce redundancy in the parameters (Boureau and Cun, 2008; Romero et al., 2015). The learned parameters are used as initialisation points in CSKNs learning (i.e., the initial value of $\theta = \{W, b\}$ of each layer). The algorithm iteratively creates a layer-specific sparse target of the input data and optimises the dictionary by minimising the error between the output of the layer and the sparse target. The degree of sparsity is therefore controlled and learned differently at each layer. The parameters of the layer are then calculated as follows:

$$\theta^l = \arg\min_{\theta^l} \|O^l - T^l\|_{\mathcal{H}}^2, \quad (7)$$

where $O^l \in \mathcal{R}^{N_b \times N_h}$ are the data vectors in RKHS, which are represented as a weighted combination of the training samples used to construct the kernel matrix at layer $l$, and $T^l$ denotes the sparse target of the layer that addresses population and lifetime sparsity.

Algorithm 1 is the pseudocode of the single layer EPLS derivation. Let us define $O_j$ as an element of row vector $O$ and denote $O^l$ as $N_b$ output vectors of dimensionality $N_h$, where $N_b$ is the size of mini-batch. Starting with no activation in $T^l$ (line 1), input patches of $O^l$ are normalised between 0 to 1 (line 2). The algorithm iteratively processes a row $O_j$ of $O^l$ by selecting the $k$th element of the $n$-th row of $O^l$ that has the maximal activation value $O_k$ minus an inhibitor $c_j$ (line 5). Here, the inhibitor is an accumulator that counts the number of times an output $j$ has been selected, increasing its inhibitor by $N_h/N$ until reaching maximal inhibition, where N is the total number of training patches. This enforces the *lifetime sparsity* and prevents the selection of an output that has already been activated $N_h/N$ times. The $k$th element of $n$-th row of target matrix $T^l$ is then activated as in line 6 (i.e., by assigning 1), considering *population sparsity*. The inhibitor is progressively updated and finally the output target is remapped to active and inactive values of corresponding non-linearity. The optimisation in relation to Equation 7 is performed using standard stochastic gradient descent (SGD) with adaptive learning rates (Schaul et al., 2013).

---

**Algorithm 1:** Single Layer EPLS

**Input**: $O$, $a$, $N$
**Output**: $T$, $c$
1: $T = 0$
2: $O = (O - \min(O))/(\max(O) - \min(O))$
3: **for** n = 1 **to** $N_b$
4:     $O_j = O_{n,j} \forall j \in \{1,2,3,\dots,N_h\}$
5:     $k = argmax_j(O_j - c_j)$
6:     $T_{n,k} = 1$
7:     $a_k = a_k + N_h/N$
8: **end for**
9: Remap $T$ to active/inactive values

---

$k$ is the output that has to be activated in the $n$-th row of $T$ and $c_j$ is an accumulator that counts the number of times an output $j$ has been selected.

### 2.2.6 Multi-layer convolutional sparse kernel networks (CSKNs)

A CSKN kernel can be learned in a hierarchical fashion for a deeper and potentially improved high-level semantic feature representation. Essentially: 1) the input feature map of layer $l+1$ can be computed by applying the convolution operation, learned weights and biases to kernel maps from layer $l$; 2) EPLS is then used to learn initial sparse features that are used as a starting point of CSKN learning; 3) a multi-layer CSKN is learned in a feedforward manner, using a given input sub-patch of size $S_z$, and kernel parameters $\alpha$ and $\beta$ for each layer.

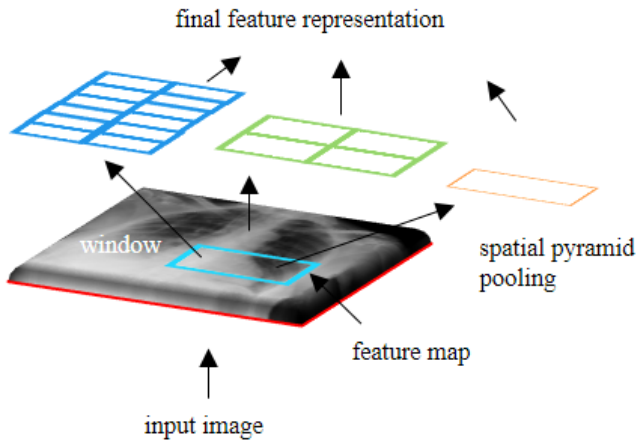

Fig 4. The SPP layer on top of CSKN.

### 2.2.7 Capturing subtle geometric variations with SPP layer

We added SPP as the last feature pooling layer to extract a final image representation that also captures subtle geometric variations. The outputs of the SPP layer are $p \cdot M$ dimensional vectors with $M$ multi-level spatial bins ($p$ is the filter size). We determined the window size of each pyramid level ($n$) based on the last feature maps ($x \times x$) generated from CSKN, as $win = x/n$. We then pooled and aggregated the responses of each filter by selecting the maximum values (max pooling) across different locations and over different spatial scales of the kernel map. This provides invariant image representations that are more robust to local transformations. Fig 2 and Fig 4 show a SPP layer combined with our CSKN.

## 3. Experimental

### 3.1 Evaluation

To evaluate our framework we compared it to other unsupervised and supervised learning methods:

a) Conventional unsupervised feature learning methods: SIFT+BoVW, Independent Component Analysis (ICA), and sparse coding (Lee et al., 2006). We implemented the SIFT descriptor together with BoVW model (SIFT+BoVW). We used a patch size of 16x16 pixels with spacing of 8 pixels in the extraction of SIFT descriptors. We used the standard codebook size of 1000 (Avni et al., 2011). The number of filters (i.e., weights) for the first layer of the ICA, and sparse coding were all set to 1600 (Romero et al., 2015).

b) State-of-the-art unsupervised learning methods: SSAE (Hinton et al., 2006; Shin et al., 2013) and CKN (Mairal et al., 2014) . The number of filters for the first layer of the SSAE was set to 1600, which was consistent with the conventional baselines above; we set the number of filters for the second layer to 1024. As the CKN is also a kernel learning method, for the purpose of comparison we trained the CKN using the same parameters as our proposed CSKN (see Section 3.2).

c) State-of-the-art supervised pre-trained CNNs (with natural images). We used the AlexNet (Krizhevsky et al., 2012), VGG (Simonyan and Zisserman, 2014), GoogLeNet (Szegedy et al., 2015), and ResNet (He et al., 2016), which have achieved high rankings in object recognition and localisation from the ImageNet Challenge. For all pre-trained CNNs models, the final fully-connected layers were used as the feature extractors.

d) A recent representative sparsity-based pre-trained CNN (with natural images) where we used a model in which 65% of the dense parameters of ResNet were removed (Liu et al., 2018). The accuracy of this pruned model was most similar to its counterpart of original pre-trained ResNet. The final fully-connected layer was used as the feature extractor.

e) State-of-the-art supervised fine-tuned CNNs. We used the same models as in the pre-trained baselines above: AlexNet (Krizhevsky et al., 2012), VGG (Simonyan and Zisserman, 2014), GoogLeNet (Szegedy et al., 2015), and ResNet (He et al., 2016). For medical image analysis, these fine-tuned CNNs have been shown to perform as well as fully trained CNNs or even outperform when there is limited training data (Kumar et al., 2017),(Tajbakhsh et al., 2016),(Shin et al., 2016). All of the models were trained for 60 epochs with the IRMA dataset. We used a batch size of 128 and an initial learning rate of $10^{-4}$. We used learning rate annealing, decaying the rate by a factor of 10 when the error plateaued.

## 3.2 Implementation details

CSKNs have four parameters that need to be determined for each layer: size of sub-patch, coefficients $\alpha$ and $\beta$, and pooling factor or filter size $p$. The parameters of our Gaussian kernel $\alpha$ and $\beta$ are automatically determined for each layer: the $\beta$ was set to be the pooling factor divided by $\sqrt{2}$; $\alpha$ was set to be the 0.1 quantile of the distribution of pair-wise distances between sub-patches, consistent with the work reported by (Mairal et al., 2014). In our experimentation, the final results were insensitive to the use of smaller quantiles such as 0.01 and 0.001. This is also consistent with other research studies, e.g., (Paulin et al., 2015).

**X-Ray image retrieval (IRMA dataset):** We adopted a two-layer architecture that was shown to perform better on gray-scale images (Paulin et al., 2015). We used the gradient map (defined in Section 2.2.3) as the input of the initial layer of our architecture; the gradient map as input has been shown to perform better than raw patches (Mairal et al., 2014). Our parameter selection process searched within a restricted space to find the optimal values of the parameters. We used values in the range 2 to 8 for sub-patch sizes and pooling factors of 100, 256, 512, 800 and 1024. For the SPP layer, we used a 4-level spatial pyramid (1x1, 2x2, 3x3, 6x6) of 50 spatial bins in all of our experiments.

**Medical image modality classification (ImageCLEF dataset) and skin disease classification (ISIC dataset):** We used the same settings as the X-ray image retrieval (described above) but used raw patches instead of gradient maps as the input because the raw patches performed better when working with RGB images. We then empirically chose the remaining parameters as shown in Table 1. For all learned features, we used the setup of the multi-class linear SVM introduced by (Yang et al., 2009), who used a differentiable quadratic hinge loss so that the training could easily be done with simple gradient-based optimisation methods. We used LBFGS with a learning rate of 0.1 and a regularization parameter of 1, consistent with the parameters specified by (Yang et al., 2009).

Table 1. FOR EACH LAYER, THE SUB-PATCH SIZE, SUB-SAMPLING FACTOR, AND THE NUMBER OF POOLING FACTOR ARE SHOWN. FOR INITIAL GRADIENT MAP, THE VALUES 16 INDICATES THE NUMBER OF ORIENTATIONS.

| Dataset | Layer | Sub-patch Size | Sub-sampling Factor | Pooling Factor |
|---|---|---|---|---|
| IRMA | Layer 1 | 1x1 | 4 | 16 |
|  | Layer 2 | 3x3 | 4 | 1024 |
| ImageCLEF and ISIC | Layer 1 | 2x2 | 2 | 100 |
|  | Layer 2 | 2x2 | 4 | 800 |

### 3.3 Computation

All the neural networks - CSKN, SSAE, CKN, fine-tuned CNNs - were trained with a GeForce GTX 1080 Ti GPU (11GB memory). It took 8 hours for our CSKN to be trained with this GPU on a machine with Intel Core i7-6800K 3.40 GHz (6 cores) processor.

### 3.4 X-ray image retrieval

We conducted medical image retrieval experiments on the IRMA dataset (Avni et al., 2011) and classification experiments on the ImageCLEF dataset (Villegas et al., 2016) and the ISIC dataset (Codella et al., 2018). For the medical image retrieval experiments, we used the ground truth annotations (i.e., IRMA code) to measure the relevance of similarity. Each test image was then used as a query image and the training images were ranked according to the Euclidean distance from the query image. For quantitative comparisons, we used precision estimates at $Q = 1, 5, 10$, and 30 as follows:

$$\text{Precision@Q} = \frac{\#\ relevant\ images\ in\ top\ Q\ images\ retrieved}{\#\ images\ of\ Q\ retrieved\ images}. \quad (8)$$

### 3.5 Medical image modality classification

For the classification experiments, we used the Top 1 accuracy (the correctness of the predicted label), which is the standard performance measure adopted in recent CNN studies for the classification of medical image modalities (Kumar et al., 2017). For the results of the supervised CNNs models with ImageCLEF dataset, we used the results reported in their respective papers.

### 3.6 Skin diseases classification

We used the area under curve (AUC) from the receiver operating characteristics (ROC) curve which were the main evaluation metrics in the ISIC 2017 competition (Codella et al., 2018). For the results of the supervised CNNs models with ISIC dataset, we used the results reported in their respective papers.

# 4. Results

The results of image retrieval experiments are shown in Table 2. We show sample results of the query and retrieval of varying structures in Fig 5. The query images are the shoulder of the scapulo-humeral joint (top row), shoulder of the acromio-clavicular joint (middle row), and (bottom left) forearm (bottom row), with artifacts including plates, screws and wires. The retrieved images are ranked by the order of similarity from left to right (top 1 to 3). Our framework had greater accuracy than other unsupervised feature learning algorithms as well as other pre-trained CNN models. Furthermore, it outperformed all the fine-tuned CNNs, achieving a top 1 precision of 52.97%. The fine-tuned GoogLeNet method achieved the best precision when considering the top 5, 10, and 30 retrieved images.

The results of image modality classification experiments are shown in Table 3. We compared our approach with several conventional unsupervised feature learning methods as well as the supervised image-based methods presented in the competition held in 2016. Our CSKN had greater accuracy than other unsupervised approaches, achieving a top 1 accuracy of 70.99%. The second best unsupervised method was SSAE with an accuracy of 65.17%. The best performing supervised method was the fine-tuned ResNet-152 with an accuracy of 85.38% (Koitka and Friedrich, 2016).

Table 4 shows the results of skin diseases classification experiments. Consistent with the modality classification experiments, we compared our approach with other unsupervised feature learning methods as well as the supervised methods presented in the competition held in 2017. Our framework had greater accuracy than other unsupervised approaches, with over 10% improvement from the second best unsupervised method (SSAE), achieving a mean AUC accuracy of 76.11%. It also had a higher accuracy than pre-trained ResNet (72.35%) and fine-tuned Inception V3 (75.00%). The fine-tuned ResNet method had the best mean AUC of 91.10%

Fig 6 shows how our initialisation with sparsity-based pre-training improves the feature representation of medical images compared to other standard pre-training methods including random initialisation and the K-mean algorithm. We also show the improvement made by the SPP. The visualisation of the learned weights from the first layer of our CSKN is shown in Fig 7. This shows that our CSKN learnt common structures such as lines and edges and also identified spatial patterns and sparse regions in the medical images. We used 400,000 image patches of size 12x12 and learned 256 filters (Olshausen and Field, 1996). The results from deeper networks are shown in Fig 8. Our experiments using 3 and 4 layer CSKN architectures did not improve performance.

Table 2. Average Image Retrieval Precision Estimates (%) at Q=1, 5, 10, and 30 (Based on the IRMA dataset). The best and second-best precisions are in bold and red respectively.

| Type | Methods/ Average Q | 1 | 5 | 10 | 30 |
|---|---|---|---|---|---|
| Unsupervised | SIFT+BoVW | 34.21 | 25.42 | 21.78 | 16.32 |
| Unsupervised | SSAE (2 layers) | 38.54 | 31.74 | 27.71 | 20.57 |
| Unsupervised | ICA | 33.92 | 26.10 | 22.42 | 16.69 |
| Unsupervised | Sparse Coding | 31.27 | 23.85 | 20.64 | 15.32 |
| Supervised | *Pre-trained* AlexNet | 37.91 | 30.46 | 26.72 | 20.90 |
| Supervised | *Pre-trained* VGG-16 | 39.29 | 32.39 | 29.25 | 24.17 |
| Supervised | *Pre-trained* VGG-19 | 38.83 | 32.46 | 29.54 | 24.20 |
| Supervised | *Pre-trained* GoogLeNet-22 | 40.39 | 33.90 | 31.09 | 26.10 |
| Supervised | *Pre-trained* sparsity-based ResNet | 40.40 | 33.95 | 30.01 | 23.60 |
| Supervised | *Pre-trained* ResNet-152 | 41.31 | 34.48 | 31.06 | 24.80 |
| Supervised | *Fine-tuned* AlexNet | 44.48 | 36.93 | 32.87 | 26.73 |
| Supervised | *Fine-tuned* VGG-16 | 48.75 | 43.73 | 40.40 | 34.59 |
| Supervised | *Fine-tuned* VGG-19 | <span style="color:red">49.45</span> | 43.94 | <span style="color:red">40.98</span> | <span style="color:red">34.87</span> |
| Supervised | *Fine-tuned* GoogLeNet | 49.39 | **44.61** | **43.12** | **38.70** |
| Supervised | *Fine-tuned* ResNet | 47.20 | 41.66 | 39.11 | 34.56 |
| **Unsupervised** | **Our CSKN** | **52.97** | <span style="color:red">44.18</span> | 39.87 | 31.59 |

Table 3. Image Classification Accuracy (%) using ImageCLEF dataset.

| Type | Methods | Accuracy (%) |
|---|---|---|
| Unsupervised | Sparse Coding | 57.08 |
| Unsupervised | ICA | 58.79 |
| Unsupervised | SSAE (2 layers) | 65.17 |
| Supervised | VGG-like CNN (500 epochs) (Semedo and Magalhães, 2016) | 65.31 |
| Unsupervised | Our CSKN | 70.99 |
| Supervised | *Pre-trained* sparsity-based ResNet (Liu et al., 2018) | 76.60 |
| Supervised | *Fine-tuned* AlexNet (100 epochs) with data augmentation (Kumar et al., 2016) | 77.55 |
| Supervised | *Modified* GoogLeNet (60 epochs) with additional data (Koitka and Friedrich, 2016) | 81.03 |
| Supervised | Ensemble of CNNs (50 epochs) with data augmentation (Kumar et al., 2017) | 82.48 |
| Supervised | *Fine-tuned* ResNet-152 with additional data (Koitka and Friedrich, 2016) | **85.38** |

Table 4. Skin Diseases Classification AUC (%) using ISIC dataset.

| Type | Methods | Seborrheic keratosis AUC | Melanoma AUC | Mean AUC |
|---|---|---|---|---|
| Unsupervised | ICA | 49.00 | 58.62 | 53.81 |
| Unsupervised | SC | 66.87 | 56.88 | 61.87 |
| Unsupervised | SSAE (2 layers) | 68.28 | 60.98 | 64.63 |
| Supervised | *Pre-trained* sparsity-based ResNet (Liu et al., 2018) | 79.60 | 60.20 | 69.90 |
| Supervised | *Pre-trained* ResNet-152 | 79.43 | 65.24 | 72.35 |
| Supervised | *Fine-tuned* Inception V3 (Murphree and Ngufor, 2017) | 81.70 | 68.40 | 75.00 |
| Unsupervised | Our CSKN | 84.85 | 67.37 | 76.11 |
| Supervised | *Pre-trained* VGG-19 | 80.20 | 73.14 | 76.67 |
| Supervised | An ensemble of CNNs (Sousa and de Moraes, 2017) | 84.00 | 80.50 | 82.30 |
| Supervised | *Fine-tuned* ResNet-101 (Bi et al., 2017) | 87.00 | **92.10** | 89.60 |
| Supervised | F*ine-tuned* ResNet-50 (Matsunaga et al., 2017) | **95.30** | 86.80 | **91.10** |

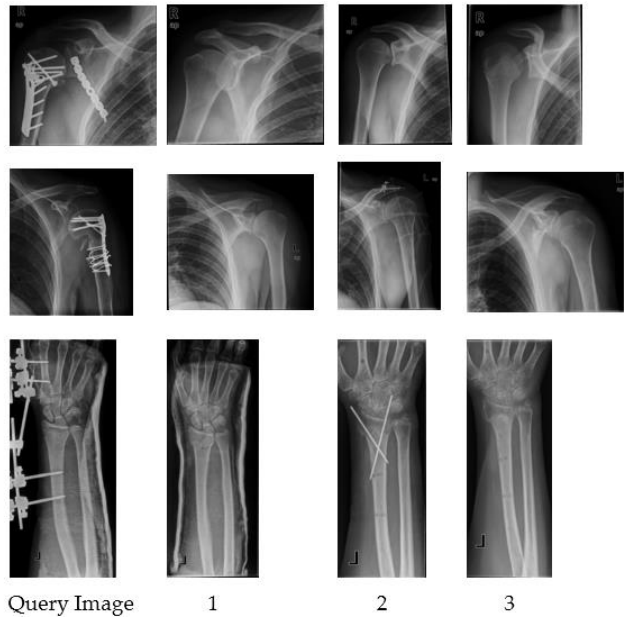

Fig 5. Sample results of query and retrieval of X-ray images using CSKN.

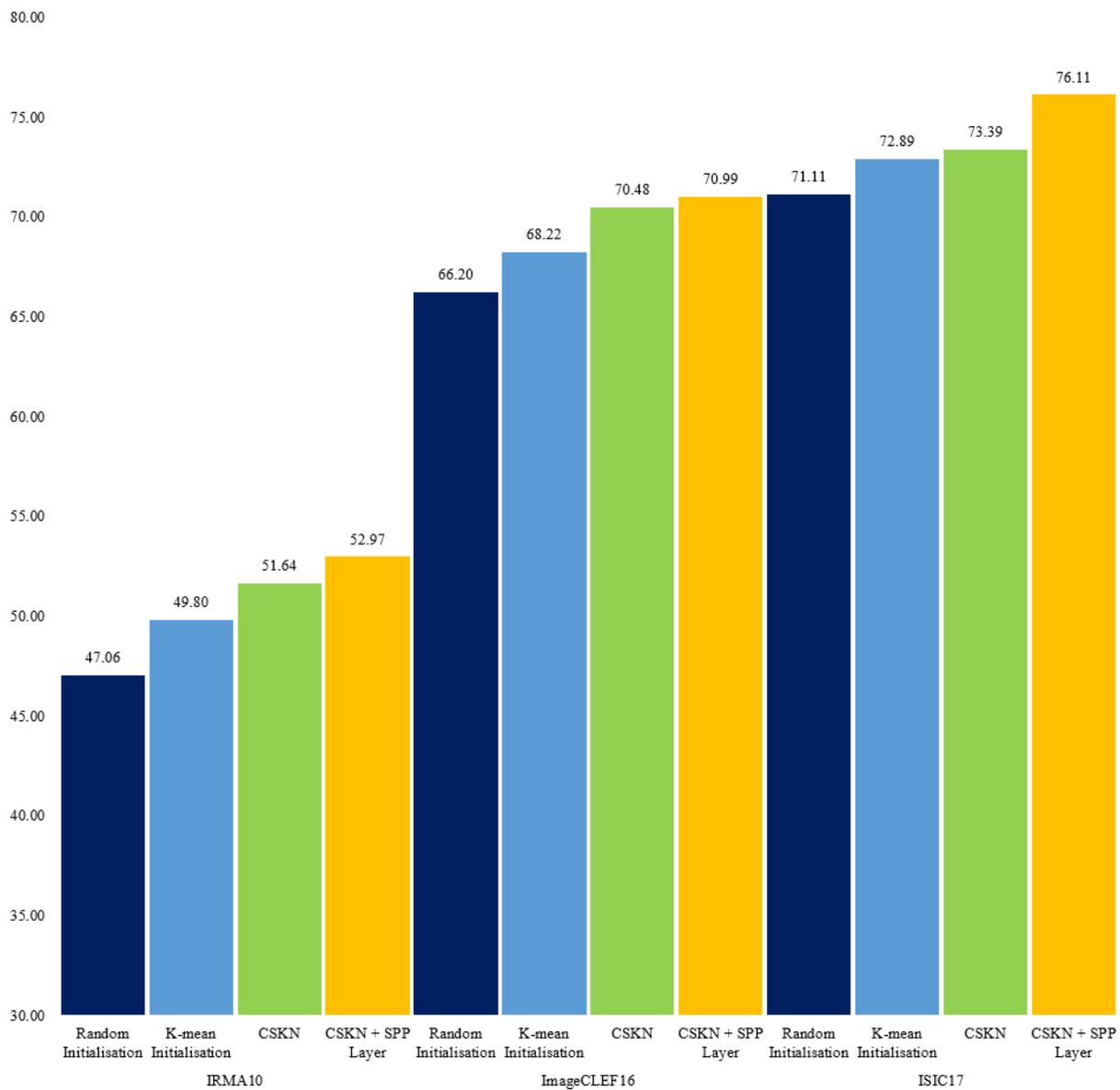

Fig 6. Top 1 average precision, accuracy and mean AUC of CKN with random and K-mean initialisation, and our Improved CSKN with SPP.

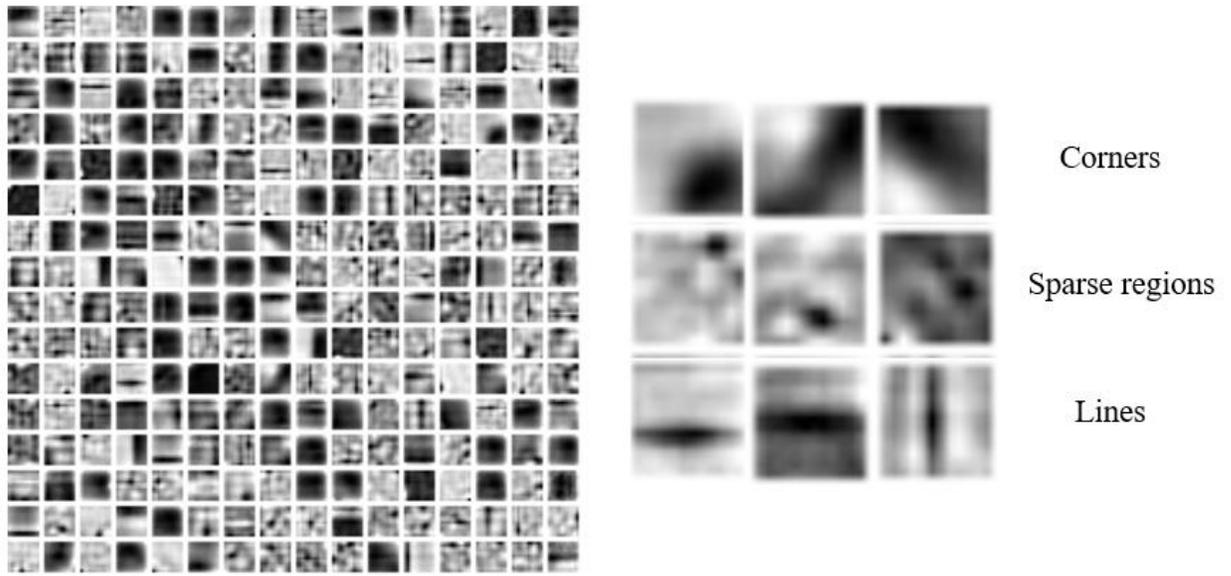

Fig 7. The visualisation of learned weights by the first layer of the CSKN using ImageCLEF dataset (gray-scale).

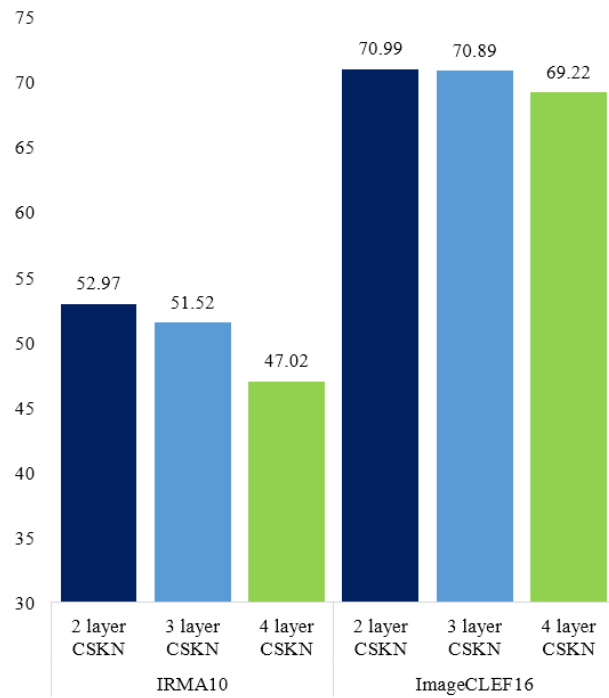

Fig 8. Results of retrieval and classification using deeper layers of CSKN.

# 5. Discussion

Our results show that our CSKN outperformed other conventional unsupervised approaches; it had comparable accuracy to state-of-the-art supervised CNNs in X-ray image retrieval and comparable accuracy to other supervised counterparts in medical image modality classification and classification of some skin conditions. Further, we showed that sparsity-based pre-training improves the feature representation of medical images, and we attribute this to our robust pre-training scheme which provided good initialisation points for subsequent convolutional kernel learning. It acts as a form of regularisation that restricts parameters to certain spaces that are more discriminative for medical image data (Erhan et al., 2010),(Mishkin and Matas, 2015). The SPP framework also improves feature representation in medical images (see Fig 6) through a multi-level spatial feature pooling technique that effectively characterises the local geometry information in the image data. To the best of our knowledge, this is the first work that couples unsupervised pre-training with unsupervised learning frameworks when compared to the conventional approache which is to combine unsupervised pre-training with subsequent supervised learning (Erhan et al., 2010),(Salimans and Kingma, 2016).

For X-ray image retrieval, our unsupervised CSKN achieved the highest accuracy (52.97%) when Q=1 (see Table 2), suggesting that the CSKN was able to learn and extract data-specific features. The quality of the features learned with conventional unsupervised hand-crafted features such as SIFT coupled with BoVW model, sparse coding, and ICA were not as robust as that of the SSAE. The accuracy of pre-trained CNNs was lower than our method as these approaches extracted features that were not tuned to a particular dataset or application, and as such have limited capacity to extract the most meaningful or discriminative features. The deeper network of pre-trained CNNs had higher accuracy (e.g., VGG-16 to ResNet-152 layers) and the fine-tuned GoogLeNet had the highest accuracy in top 5, 10, 30. We attribute this to its network architecture exploiting the local sparse structure of a convolutional network (Szegedy et al., 2015). Our method was designed to learn class-specific image features for better discrimination in an unsupervised fashion but this means it can be sensitive to subtle inter-class variations which is why accuracy drops at a higher rate compared to supervised counterparts as more subtly similar images are retrieved. For medical image retrieval applications, the five most similar images (i.e., top 5) for a query are commonly used for comparative analysis (Quellec et al., 2010). Our CSKN achieved a competitive top 5 accuracy (44.18%), which was the second best after the fine-tuned GoogLeNet (44.61%).

In medical image modality classification, our unsupervised CSKN outperformed all other unsupervised approaches and achieved a comparable accuracy to all supervised CNNs that were part of the ImageCLEF 2016 challenge. Similar to the X-ray image results, the quality of image features extracted using conventional unsupervised approaches, sparse coding and ICA, were not as robust as that of the SSAE. Unlike sparse coding and ICA, SSAE learned image features in a hierarchal manner and hence was the closest method to our approach. The top performing methods were all based on well-established supervised CNNs including AlexNet (Kumar et al., 2016), VGG (Semedo and Magalhães, 2016), GoogLeNet (Koitka and Friedrich, 2016), and ResNet (Koitka and Friedrich, 2016). These CNNs were trained from scratch or fine-tuned with medical images to derive high-level data specific features. As expected, deeper CNNs also had higher accuracy than shallower CNNs (see Table 3). Our unsupervised CSKN (accuracy of 70.99%) performed better than supervised VGG-like CNNs (65.31%) (Semedo and Magalhães, 2016) with over 5% improvement in modality classification. While most of referenced methods used the same training data, the method reported by (Koitka and Friedrich, 2016) that had the best performance in the competition, added extra data from additional sources which contributed to its overall accuracy.

The ImageCLEF dataset also contains different generic biomedical illustrations such as gene sequences or chemical structures, and so in comparison to the X-ray IRMA dataset, there were more diverse and complex variations in image characteristics. As a consequence, the overall performance of our CSKN compared to other supervised approaches was poorer on the ImageCLEF dataset than the IRMA dataset. Nevertheless, our method was able to derive discriminative medical image features from a variety of image modalities without reliance on labels, and its accuracy was better than that of supervised VGG-like CNNs (Semedo and Magalhães, 2016).

In classification of the skin lesions our unsupervised CSKN outperformed all other unsupervised feature learning methods and achieved a higher accuracy (76.11%) than pre-trained ResNet (72.35%) and fine-tuned Inception V3 (75.00%). Consistent to the results from other datasets, SSAE was the next best approach among other unsupervised methods. The top performing approaches reported in the competition also used fine-tuned CNNs, e.g., AlexNet, VGG, Inception v3, and ResNet. The best performing methods (Matsunaga et al., 2017) (Bi et al., 2017), however, added extra data from additional sources. This indicates that the fine-tuned CNNs are still dependent on the availability of labelled data. Our CSKN, on the other hand, was able to learn meaningful medical image features in a completely unsupervised fashion.

## 5.1 Limitations and Future work

Although our approach learned medical image feature representations without supervision and labels, some of the parameters (including sub-patch size, sub-sampling factor, or pooling factor (i.e., filter size) for each layer, must be empirically derived (see Section 3.2). Generally, smaller subsampling factors and larger pooling factors led to better performance at the cost of increased computational complexity. Nevertheless, our results show that sparsity-based pre-training and SPP pooling consistently improved overall feature representation even when different parameters were used. We used an integral form of the Gaussian Radial Basis Function (RBF) kernel to approximate the kernel map (image feature representation in a RKHS). A multi-kernel approach (Song et al., 2018) describing diverse properties of medical images could potentially provide more meaningful feature representation and we will explore such approaches in the future.

We suggest that our unsupervised initialisation will benefit supervised learning approaches when there are limited labelled training data. When CSKN is used to initialise a CNN for supervised fine-tuning, it could potentially enable the derivation of semantically more meaningful representations of the image data than traditional CNN fine-tuning approaches that are initialised with natural images. The investigation of the impact on fine-tuning is a substantial research study in itself and so we will pursue this in future work. Since our CSKN is completely unsupervised, we suggest that it can be considered as an important first step to accessing the large volume of unannotated data in medical imaging repositories. We note that compared to other supervised CNNs, our CSKN requires learning fewer parameters across fewer layers (two layers in this paper), and therefore, can be efficiently coupled with subsequent supervised learning approaches without a large computational cost.

## 6. Conclusion

We have proposed a new unsupervised sparsity-based feature learning framework for characterisation of medical image data. Our layerwise pre-training, using convolutional sparse features, improved the learning outcomes and feature representations in image retrieval and classification. We compared our approach to other unsupervised and supervised methods on three large public datasets and showed that our approach was competitive with the state-of-the-art supervised CNNs. Our approach demonstrated the feasibility of using large collections of unlabelled medical data to characterise medical image features and offers the opportunity to access the large volume of unannotated data that are available in medical imaging repositories.